\documentclass[journal]{IEEEtran}
\usepackage{url}
\usepackage{epsfig}
\usepackage{epstopdf}
\usepackage{cite}
\usepackage{caption}

\usepackage{times}
\usepackage{epsfig}
\usepackage{graphicx}
\usepackage{amsmath}
\usepackage{amssymb}
\usepackage{rotating}
\usepackage{array}

\ifCLASSINFOpdf
\else
\fi

\begin{document}
\title{What is a salient object? A dataset and \\ a baseline model for salient object detection}


\author{Ali~Borji, \IEEEmembership{Member,~IEEE}
        \IEEEcompsocitemizethanks{
\IEEEcompsocthanksitem A. Borji is with the Computer Science Department, University of Wisconsin, Milwaukee, WI, 53211. E-mail: borji@uwm.edu. \protect 
\IEEEcompsocthanksitem Manuscript received May 2014. }
\thanks{}}

%
%

\markboth{IEEE TRANSACTIONS ON IMAGE PROCESSING,~Vol.XXX, No.XXX, XXXXX~2014}%
{Shell \MakeLowercase{\textit{{\em et al.}}}: xx IEEE TRANSACTIONS ON IMAGE PROCESSING}

\maketitle

\begin{abstract}
 
Salient object detection or salient region detection models, diverging from fixation prediction models, have traditionally been dealing with locating and segmenting the most salient object or region in a scene.
While the notion of most salient object is sensible when multiple objects exist in a scene, current datasets 
for evaluation of saliency detection approaches often have scenes with only one single object. We introduce three main contributions in this paper: 
First, we take an in-depth look at the problem of salient object detection by studying the relationship between where people look in scenes and what they choose as the most salient object when they are explicitly asked. Based on the agreement between fixations and saliency judgments, we then suggest that the most salient object is the one that attracts the highest fraction of fixations.
Second, we provide two new less biased benchmark datasets containing scenes with multiple objects that challenge existing saliency models. Indeed, we observed a severe drop in performance of 8 state-of-the-art models on our datasets (40\% to 70\%). Third, we propose a very simple yet powerful model based on superpixels to be used as a baseline for model evaluation and comparison. While on par with the best models on MSRA-5K dataset, our model wins over other models on our data highlighting a serious drawback of existing models, which is convoluting the processes of locating the most salient object and its segmentation.
We also provide a review and statistical analysis of some labeled scene datasets that can be used for evaluating salient object detection models. We believe that our work can greatly help remedy the over-fitting of models to existing biased datasets and opens new venues for future research in this fast-evolving field. 

\end{abstract}

 

\section{Introduction}

\begin{figure}[t]

\centering	
\includegraphics[width=8.5cm,height=7cm ]{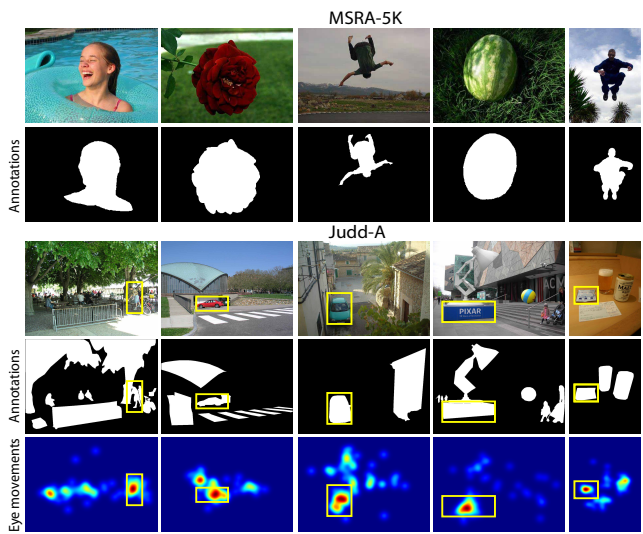} 

\caption{Top: sample images from the MSRA-5K dataset~\cite{Jiang}. Bottom: sample images from Judd-A (Judd-Annotation) dataset~\cite{Judd2009} with fixations and our annotations. Yellow boxes contain peaks of fixation maps. While scenes in MSRA-5K~\cite{JiangCVPR2013} (and other salient object datasets; Table 1) have close-up views of single objects in uniform and structured backgrounds, scenes in Judd-A dataset are more cluttered and have larger fields of view.}
\label{fig:Fig1}
\end{figure}

\IEEEPARstart{P}LEASE take a look at the images in the top row of Fig.~\ref{fig:Fig1}. Which object stands out the most (i.e., is the most salient one) in each of these scenes?
The answer is trivial. There is only one object, thus it is the most salient one. Now, look at the images in the third row. 
These scenes are much more complex and contain several objects, thus it is more challenging for a vision system to select the most salient object.

This problem, known as \textit{salient object detection (and segmentation)}, has recently attracted a great deal of interest in computer vision community.
The goal is to simulate the astonishing capability of human attention in prioritizing objects for high-level processing. 
Such a capability has several applications in recognition (e.g.,~\cite{rutishauser2004bottom,Uijlings13,borji2011scene}), image and video compression (e.g.,~\cite{guo2010novel,itti2004automatic}), video summarization (e.g.,~\cite{ma2005generic,Bodesheim2011}, media re-targeting and photo collage (e.g.,~\cite{goferman2010puzzle,wang2006picture}), image quality assessment (e.g.,~\cite{ninassi2007does,liu2009studying}), image segmentation (e.g.,~\cite{donoser2009saliency}), content-based image retrieval and image collection browsing (e.g.,~\cite{chen2009sketch2photo,feng2010attention}), image editing and manipulating (e.g.,~\cite{chia2011semantic,liu2012web}), visual tracking (e.g.,~\cite{stalder2013dynamic,li2013visual,BorjiTracking}), object discovery (e.g.,~\cite{karpathyobject,frintropcognitive}), and human-robot interaction (e.g.,~\cite{meger2008curious}).


A large number of saliency detection methods have been proposed in the past 7 years (since~\cite{LiuSunZhengTangShumCVPR2007}). 
In general, a salient object detection model involves two steps: 1) \textit{selecting objects to process} (i.e., determining saliency order of objects), and 2) \textit{segmenting the object area} (i.e., isolating the object and its boundary). So far, models have bypassed the first challenge by focusing on scenes with single objects (See Fig.~\ref{fig:Fig1}). They do a decent job on the second step as witnessed by very high performances on existing biased datasets (e.g., on ASD dataset~\cite{AchantaCVPR}) which contain low-clutter images with often a single object at the center. However, it is unclear how current models perform on complex cluttered scenes with several objects. Despite the volume of past research, this trend has not been yet fully pursued, mainly due to the lack of two ingredients: 1) suitable benchmark datasets for scaling up models and model development, and 2) a widely-agreed objective definition of the most salient object.  In this paper, we strive to provide solutions for these problems. Further, we aim to discover which component might be the weakest link in the possible failure of models when migrating to complex scenes. 


Some related topics, closely or remotely, to visual saliency modeling and salient object detection include: object importance~\cite{spain2011measuring,berg2012understanding}, object proposal generation~\cite{Alexe},  memorability~\cite{isola2011makes}, scene clutter~\cite{rosenholtz2007measuring}, image interestingness~\cite{katti2008pre,gygli2013interestingness,dhar2011high}, video interestingness~\cite{jiang2013understanding}, surprise~\cite{itti2005bayesian}, image quality assessment~\cite{wang2004image}, scene typicality~\cite{vogel2004semantic,ehinger2011estimating}, aesthetic~\cite{dhar2011high}, and attributes~\cite{farhadi2009describing,Parikh}.




\section{Related work}

One of the earliest models, which generated the \textit{first wave} of interest in image saliency in computer vision and neuroscience communities, was proposed by Itti {\em et al.}~\cite{Itti98}. This model was an implementation of earlier general computational frameworks and psychological theories of bottom-up attention based on center-surround mechanisms. In~\cite{Itti98}, Itti {\em et al.} showed examples where their model was able to detect spatial discontinuities in scenes. Subsequent behavioral (e.g.,~\cite{ParkhustEtal2002}) and computational studies (e.g.,~\cite{BruceTsotsos2005}) started to predict fixations with saliency models to verify models and to understand human visual attention.
A \textit{second wave} of interest appeared with works of Liu {\em et al.}~\cite{LiuSunZhengTangShumCVPR2007} and Achanta {\em et al.}~\cite{AchantaCVPR} who treated saliency detection as a binary segmentation problem with 1 for a foreground pixel and 0 for a pixel of the background region. Since then it has been less clear where this new definition stands as it shares many concepts with other well-established computer vision areas such as general segmentation algorithms (e.g.,~\cite{Arbelaez_etal11pami,Martin_etal04pami}), category independent object proposals (e.g.,~\cite{Alexe}), fixation prediction saliency models (e.g.~\cite{BruceTsotsos2005,Judd2009,HouZhangNIPS2008}), and general object detection methods. This is partly because current datasets have shaped a definition for this problem, which might not totally reflect full potential of models to \textit{select and segment salient objects in an image with an arbitrary level of complexity}.




Reviewing all saliency detection models goes beyond the scope of this paper (See~\cite{BorjiSalRev,BorjiECCV,BorjiTIP,BorjiICCV2013}). Some breakthrough efforts are as follows. Liu {\em et al.}~\cite{LiuSunZhengTangShumCVPR2007} introduced a conditional random field (CRF) framework to combine multi-scale contrast and local contrast based on surrounding, context, and color spatial distributions for binary saliency estimation. 
Achanta {\em et al.}~\cite{AchantaCVPR} proposed subtracting the average color from the low-pass filtered input for saliency detection. 
Goferman {\em et al.}~\cite{Goferman} used a patch-based approach to incorporate global context, aiming to detect image regions that represent the scene.
Cheng {\em et al.}~\cite{Chang} proposed a region contrast-based method to measure global contrast in the Lab color space. 
In~\cite{WangCVPR2011}, Wang {\em et al.} estimated local saliency, leveraging a dictionary learned from other images, and global saliency using a dictionary learned from other patches of the same image.
Perazzi {\em et al.}~\cite{Perazzi} observed that decomposing an image into perceptually uniform regions, which abstracts away unnecessary details, is important for high quality saliency detection. 
In~\cite{Jiang}, Jiang {\em et al.} utilized the difference between the color histogram of a region and its immediately neighboring regions for measuring saliency.
Feng {\em et al.}~\cite{Feng} defined a measure of saliency as the cost of composing an image window using the remaining parts of the image, and tested it on PASCAL VOC dataset~\cite{Everingham}.
This method, in its essence, follows the same goal as in~\cite{Alexe}. Chang {\em et al.}~\cite{Chang} proposed a graphical model for fusing generic objectness~\cite{Alexe} and visual saliency for salient object detection.
Shen and Wu~\cite{Shen} modeled an image as a low-rank matrix (background) plus sparse noises (salient regions) in a feature space.
More recently, Margolin {\em et al.}~\cite{Margolin} integrated pattern and color distinctnesses with high-level cues to measure saliency of an image patch.



Some studies have considered the relationship between fixations and salincy judgments similar to~\cite{borji2013stands}. For example, Xu et al.~\cite{xu2014predicting} investigated the role of high-level semantic knowledge (e.g., object operability, watchability, gaze direction) and object information (e.g., object center-bias) for fixation prediction in free viewing of natural scenes. They constructed a large dataset\footnote{http://www.ece.nus.edu.sg/stfpage/eleqiz/predicting.html} called  ``Object and Semantic Images and Eye-tracking (OSIE)''. Indeed they found an added value for this information for fixation prediction and proposed a regression model (to find combination weights for different cues) that improves fixation prediction performance. 
Koehler et al.~\cite{koehler2014saliency} collected a dataset known as the UCSB dataset\footnote{https://labs.psych.ucsb.edu/eckstein/miguel/research\_pages/ \\ saliencydata.html}. 
This dataset contains 800 images. One hundred observers performed an explicit saliency judgment task, 22 observers performed a free viewing task, 20 observers performed a saliency search task, and 38 observers performed a cued object search task. Observers completing the free viewing task were instructed to freely view the images.
In the explicit saliency judgment task, observers were instructed to view a picture on a computer monitor and click on the object or area in the image that was most salient to them. \emph{Salient} was
explained to observers as something that stood out or caught their eye (similar to~\cite{borji2013stands}). 
Observers in the saliency search task were instructed to determine whether or not the most salient object or location in an image was on the left or right half of the scene. Finally, observers who performed the cued object search task were asked to determine whether or not a target object was present in the image. Then, they conducted a benchmark and introduced models that perform the best on each of these tasks.

A similar line of work to ours in this paper has been proposed by Mishra {\em et al.}~\cite{Mishra} where they combined monocular cues (color, intensity, and texture) with stereo and motion features to segment a region given an initial user-specified seed point, practically ignoring the first stage in saliency detection (which we address here by automatically generating a seed point). Ultimately, our attempt in this work is to bridge the interactive segmentation algorithms (e.g.,~\cite{Mishra,Rother}) and saliency detection models and help transcend their 
applicability.

Perhaps the most similar work to ours has been published by Li {\em et al.}~\cite{YiHou2014}. In their work, they offer two contributions. \textit{First}, they collect an eye movement dataset using annotated images from the PASCAL dataset~\cite{Everingham} and call their dataset PASCAL-S. \textit{Second}, they propose a model that outperforms other state-of-the-art salient object detection models on this dataset (as well as four other benchmark datasets). Their model decouples the salient object detection problem into two processes: 1) \textit{a segment generation process}, followed by 2) \textit{a saliency scoring mechanism} using fixation prediction. 
Here, similar to Li {\em et al.}, we also take advantage of eye movements to measure object saliency but instead of first fully segmenting the scene, we perform a shallow segmentation using superpixels. We then only focus on segmenting the object that is most likely to attract attention. In other words, the two steps are similar to Li {\em et al.} but are performed in the reverse order. This can potentially lead to better efficiency as the first expensive segmentation part is now only an approximation. 

We also offer another dataset which is complimentary to Li {\em et al.}'s dataset and together both datasets (and models) could hopefully lead to a paradigm shift in the salient object detection field to avoid using simple biased datasets. Further, we situate this field among other similar fields such as general object detection and segmentation, objectness proposal generation models, and saliency models for fixation prediction.

\begin{table}[t]
\renewcommand{\arraystretch}{1}
\renewcommand{\tabcolsep}{0.5mm}
\begin{center}\mbox{\footnotesize
\begin{tabular}{|l|l|l|c|l|l|c|c|}
\hline
Dataset & Ref & Total           & Num. of  &  Annt. &  Scene & Num             & Eye     \\
              &         &  Scenes     & Objects  &             &  Resolution    & Annt. & Data \\ \hline\hline

ASD & \cite{AchantaCVPR,LiuSunZhengTangShumCVPR2007} & 1000 & $\sim$1 & BD & 400 $\times$ 300 & 1 & -  \\
MSRA-A & \cite{LiuSunZhengTangShumCVPR2007} & 20K & $\sim$1 & BB & 400 $\times$ 300 & 3 & -  \\
MSRA-B & \cite{LiuSunZhengTangShumCVPR2007} & 5K & $\sim$1 & BB & 400 $\times$ 300 & 9 & -  \\
MSRA-5K & \cite{JiangCVPR2013,LiuSunZhengTangShumCVPR2007} & 5K & $\sim$1 & BD & 400 $\times$ 300 & 1 & -  \\
SED-1 & \cite{Alpert} & 100 & 1 & BD & $\sim$300 $\times$ 225 & 3 & -  \\
SED-2 & \cite{Alpert} & 100 & 2 & BD & $\sim$300 $\times$ 225 & 3 & -  \\
SOD & \cite{Movahhedi,Arbelaez_etal11pami} & 300 & $\sim$3 & BD & 481 $\times$ 321 & 7 & -  \\
CSSD & \cite{Yan} & 200 & $\sim$1 & BD & $\sim$400 $\times$ 300 & 1 & -  \\
ECSSD & \cite{Yan} & 10K & $\sim$1 & BD & $\sim$400 $\times$ 300 & 1 & -  \\
ImgSal & \cite{JianLi} & 235 & $\sim$ 2 & BD & 640 $\times$ 480 & 19 & 50  \\
THUR10K & \cite{MingCheng,LiuSunZhengTangShumCVPR2007} & 10K & $\sim$1 & BD & 400 $\times$ 300 & 1 & -  \\
THUR15K & \cite{MingCheng} & 15K & $\sim$1 & BD & 400 $\times$ 300 & 1 & -  \\
iCoseg & \cite{Batra} & 643 & $\sim$1 & BD & $\sim$500 $\times$ 400 & 1 & -  \\
DUT-OMRON & \cite{yang2013saliency} &  5,172 & $\sim$5 & BD & 400 $\times$ 400 & 5 & 5  \\
PASCAL-S & \cite{YiHou2014,Everingham} & 850 & $\sim$5 & BD & Variable & 12 & 8  \\
UCSB & \cite{koehler2014saliency} &  700 & $\sim$5 & BD & 405 $\times$ 405 & 100 & 22 \\
OSIE & \cite{xu2014predicting} & 700 & $\sim$7 & BD & 800 $\times$ 600 & 1 & 15 \\

\hline 

Bruce-A & \cite{BruceTsotsos2005} & 120 & $\sim$4 & BD & 681 $\times$ 511 & 70 & 20  \\
Judd-A & \cite{Judd2009} & 900 & $\sim$5 & BD & 1024 $\times$ 768 & 2 & 15  \\

\hline
\end{tabular}
}\end{center}

\caption{Overview of popular salient object datasets. The last two proposed here (A stands for ``Annotation'') avoid the dreaded entry of ``1'' in the number of objects. 
Compared with other datasets, scenes in Judd-A and Bruce-A datasets have more variety and 
are less structured. BB and BD stand for bounding box and boundary (i.e., pixel accuracy), respectively. 
Last column shows the number of eye tracking subjects. 
Datasets derived from the MSRA carry its problems, which are images with single objects, low clutter, and high degree of center-bias. iCoSeg is a co-segmentation dataset. 
The ASD dataset is also known as MSRA1000.}
\label{tab:db}

\end{table}


Several salient object detection datasets have been created as more models have been introduced in the literature to extend capabilities of models to more complex scenes. 
Table~\ref{tab:db} lists properties of 19 popular salient object detection datasets. Although these datasets suffer from some biases (e.g., low scene clutter, center-bias, uniform backgrounds, and non-ambiguous objects), they have been very influential for the past progress. Unfortunately, recent efforts to extend existing datasets have only increased the number of images without really addressing core issues specifically background clutter and number of objects. Majority of datasets (in particular large scale ones such as those derived from the MSRA dataset) have scenes with often one object which is usually located at the 
image center. This has made model evaluation challenging since some high-performing models that emphasize image center fail in detecting and segmenting the most salient off-center object~\cite{BorjiECCV}. We believe that now is the time to move on to more versatile datasets and remedy biases in salient object datasets.

\section{What is a salient object?}

In this section, we briefly explain how salient object detection models differ from fixation prediction models, what people consider the most salient object when they are explicitly asked to choose one, what are the relationships between these judgments and eye movements, and what salient object detection models actually predict.

We investigate properties of salient objects from humans' point of view when they are explicitly asked to choose such objects. We then study whether (and to what extent)
saliency judgments agree with eye movements. While it has been assumed that eye movements are indicators of salient objects, so far few studies (e.g.,~\cite{borji2013stands,masciocchi2009everyone}) have directly and quantitatively confirmed this assumption. Moreover, the level of agreement and cases of disagreement between fixations and saliency judgments have not been fully explored. Some studies (e.g.,~\cite{Elazary_Itti08jov}), have shown that human observers choose to annotate salient objects or regions first but they have not asked humans explicitly (LabelMe data was analyzed in~\cite{Elazary_Itti08jov}) and they have ignored eye movements. Knowing which objects humans consider as salient is specially crucial when outputs of a model are going to be interpreted by humans. 

\subsection{Salient object detection vs. fixation prediction}
There are two major differences between models defining saliency as ``where people look'' and models defining saliency as ``which objects stand out''. 
\textit{First}, the former models aim to predict points that people look in free-viewing of natural scenes usually for 3 to 5 seconds while the latter aim to detect and segment salient objects (by drawing pixel-accurate silhouettes around them).
In principle a model that scores well on one problem
should not score very well on the other. An optimal model for fixation prediction should only highlight those points that a viewer will look at (few points inside an object and not the whole object region).
Since salient object detection models aim to segment the whole object region they will generate a lot of false positives (these points belong to the object but viewers may not fixate at them) when it comes to fixation prediction. On the contrary, a fixation prediction model will miss a lot of points inside the object (i.e., false negatives) when it comes to segmentation. 

\textit{Second}, due to noise in eye tracking or observers' saccade landing (typically around 1 degrees and $\sim$ 30 pixels), highly accurate pixel-level
prediction maps are less desired. In fact, due to these noises, sometimes blurring prediction maps increases the scores~\cite{BorjiLGS,HouPAMI,YiHou2014}. On the contrary, producing salient object detection maps that can accurately distinguish object boundaries are highly desirable specially in applications. Due to these, different evaluation and benchmarks have been developed for comparing models in these two categories.

In practice, models, whether they address segmentation or fixation prediction,
are applicable interchangeably as both entail generating similar saliency maps.
For example, several researches have been thresholding saliency maps of their models, originally designed to predict fixations, to detect and segment salient proto-objects
(e.g.,~\cite{seo2009static,erdem2013visual}).
 

\begin{figure}[t]

\centering	
\includegraphics[width=\linewidth]{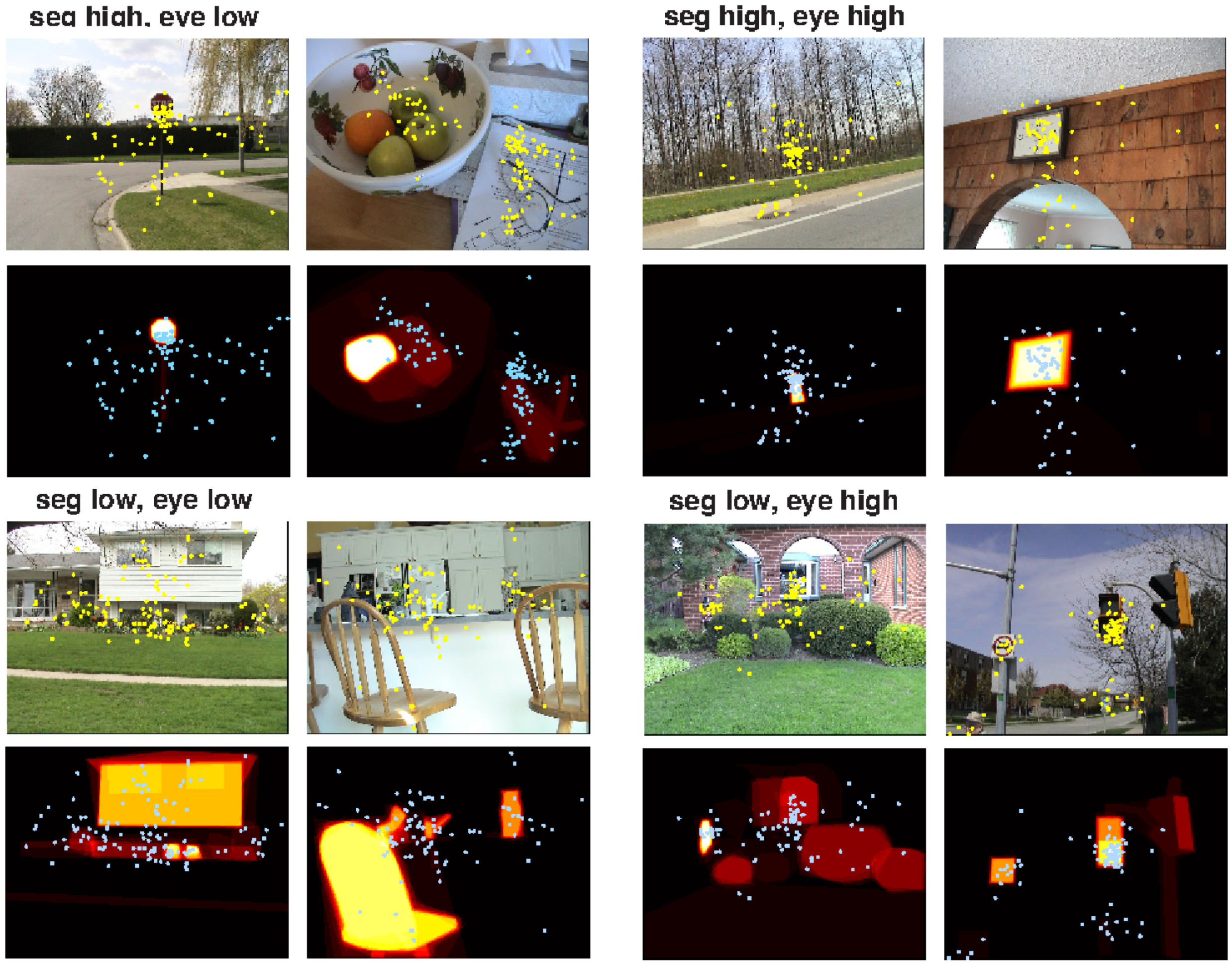} 

\caption{Sample images from Bruce-A dataset~\cite{BruceTsotsos2005} and annotation heat maps embedded with fixations. 
eye-low/eye-high indicate low/high agreement in eye movements measured in termed of shuffled AUC score~\cite{BorjiTIP}. seg-low/seg-high indicate low/high agreement in 
saliency annotations using the equation in section III.B.}
\label{fig:sampleIO-AM}
\end{figure}

\subsection{Human explicit saliency judgments}


In our previous study~\cite{borji2013stands}, we addressed what people consider as the most outstanding (i.e., salient) object in a scene. While in ~\cite{borji2013stands} we studied the explicit saliency problem from a behavioral perspective, here we are mainly interested in constructing computational models for automatic salient object detection in arbitrary visual scenes. A total of 70 students (13 male, 57 female) undergraduate USC students with normal or corrected-to-normal vision in the age range between 18 and 23 (mean = 19.7, std = 1.4)
were asked to draw a polygon around the object that stood out the most. Participants' annotations were supposed not to be too loose (general) or too tight (specific) around the object. They were shown an illustrative example for this purpose. Participants were able to relocate their drawn polygon from one object to another or modify its outline. We were concerned with the case of selection of the single most salient object in an image. Stimuli were the images from the dataset by Bruce and Tsotsos (2005)~\cite{BruceTsotsos2005}\footnote{This dataset contains eye movements over 120 color photographs of indoor and outdoor environments with the resolution of 511 $\times$ 681 pixels. Images in this dataset have been presented at random to 20 observers (in a free-viewing task) for 4 sec each, with 2 sec of delay (a gray mask) in between.}. See Fig.~\ref{fig:sampleIO-AM} for sample images from this dataset. %






 We first measured the degree to which annotations of participants agree with each other using the following quantitative measure: 
\begin{equation}
 r_{k} = \frac{2}{n(n-1)} \sum_{i=1}^{n-1} \sum_{j=i+1}^{n=70} |s_{ik} \cap s_{jk}|/|s_{ik} \cup s_{jk}|
\end{equation}

 where $s_{ik}$ and $s_{jk}$ are annotations of $i$-th and $j$-th participants, respectively (out of $n=70$ participants) over the $k$-th image. Above measure has the well-defined lower-bound of 0, when there is no overlap in segmentations of users, and the upper-bound of 1, when segmentations have perfect overlap. Fig.~\ref{fig:exp1}.left shows histogram of $r$ values. Participants had moderate agreement with each other (mean $r=0.37$; std $0.17$; significantly above chance). Inspection of images with lowest $r$ values shows that these scenes had several foreground objects while images with highest annotation agreement had often one visually distinct salient object (e.g., a sign, a person, or an animal; see Fig.~\ref{fig:sampleIO-AM}). 


\subsection{Relationship between saliency judgments and fixations}
\begin{figure}[htbp!]

\centering	
\includegraphics[width=8.5cm,height=3.9cm ]{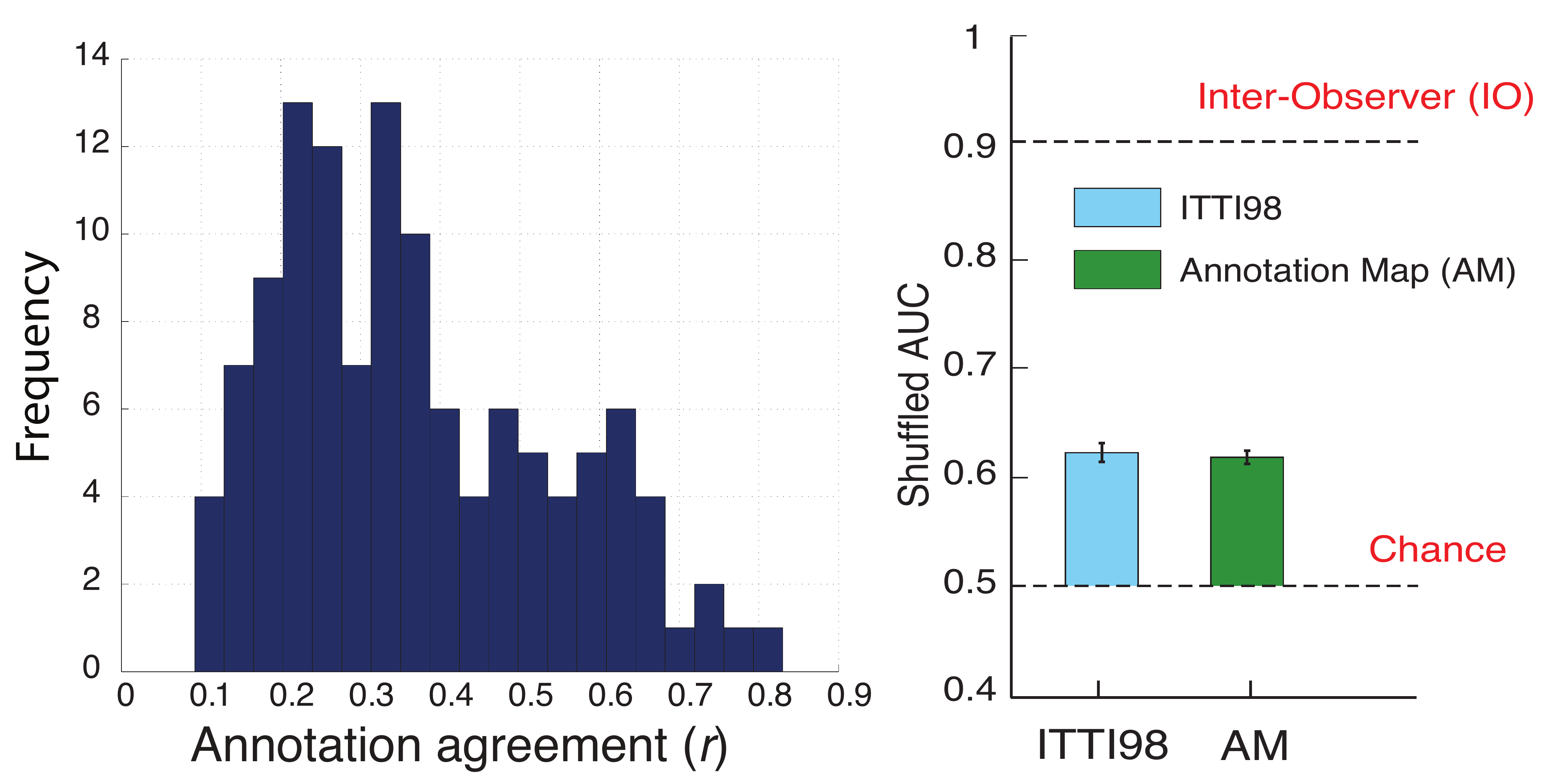} 
\caption{Left: Histogram of agreements in our explicit saliency judgment task, Right: Prediction power of the annotation map (AM) and the saliency model by Itti {\em et al.}~\cite{Itti98} for explaining eye movements over Bruce and Tsotsos dataset. Chance level is the accuracy of a random map with the value of each pixel drawn uniformly random between 0 and 1. Inter-observer (IO) model is a map build from fixations of other observers over the same image and is then smoothed with a small Gaussian kernel.}
  
\label{fig:exp1}
\end{figure}

We also investigated the relationship between explicit saliency judgments and freeviewing fixations as two indicators of visual attention. 
Here we used shuffled AUC (sAUC) score to tackle center-bias in eye movement data~\cite{CenterBiasedTatler,BorjiTIP}.
For each of 120 images, we showed that a map built from annotations of 70 participants explains fixations of free viewing observers significantly above chance (sAUC of $0.62\pm0.07$, chance $0.50$, $t$-test $p<0.05$; Fig.~\ref{fig:exp1}.right). The prediction power of this map was as good as the ITTI98 model~\cite{Itti98}. Hence, we concluded that explicit saliency judgments agree with fixations. Fig.~\ref{fig:sampleIO-AM} shows high- and low-agreement cases between fixations and annotations.



Here, we merge annotations of all 70 participants on each image, normalize the resultant map to [0 1], and threshold it at 0.7 to build our first benchmark saliency detection dataset (called Bruce-A).
Prevalent objects in Bruce and Tsotsos dataset are man-made home supplies in indoor scenes (see~\cite{BruceTsotsos2005} for more details on this dataset).

Similar results, to link fixations with salient objects, have been reported by Koehler {\em et al.}~\cite{koehler2014saliency}. As in~\cite{masciocchi2009everyone}, they asked observers to click on salient locations in natural scenes. They showed high correlation between clicked locations and observers' eye movements (from a different group of subjects) in free-viewing. While the most salient~\cite{borji2013stands}, important~\cite{spain2011measuring}, or interesting~\cite{masciocchi2009everyone,Elazary_Itti08jov,jiang2013understanding} object may tell us a lot about a scene, eventually there is a subset of objects that can minimally describe a scene. This has been addressed in the past somewhat indirectly in the contexts of saliency~\cite{Goferman}, language and attention~\cite{itti2006attention}, and phrasal recognition~\cite{kulkarni2011baby,farhadi2009describing}.

\begin{figure}[t]

\centering	
\fbox{\includegraphics[width=7.5cm,height=6cm ]{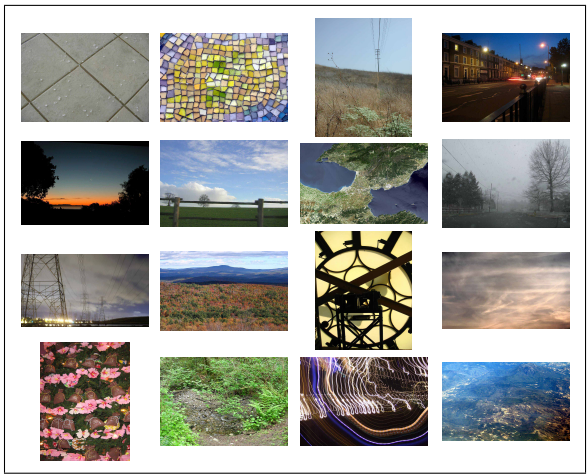}} 
\caption{Sixteen sample images from the Judd dataset that are not included in Judd-A dataset.
These images do not have well-defined salient objects in them, contain lots of background clutter
and are often boring!.}
\label{fig:FigDiscarded}
\end{figure}
 
\section{Our new large-scale dataset: Judd-A}
 
Based on results from our saliency judgment experiment~\cite{borji2013stands}, we then decided to annotate scenes of the dataset by Judd {\em et al.}~\cite{Judd2009}.
The reason for choosing this dataset is because it is currently the most popular dataset for benchmarking fixation prediction models~\cite{Judd2009,BorjiTIP}.
It contains eye movements of 15 observers freely viewing 1003 scenes from variety of topics. Thus, using fixations we can easily determine which object, out of several annotated objects, is the most salient one. 
We only used 900 images from the Judd dataset and discarded images without well-defined objects (e.g., mosaic tiles, flames) or images with very cluttered backgrounds (e.g., nature scenes).
Figure~\ref{fig:FigDiscarded} shows examples of discarded scenes.

\begin{figure*}[htp!]

\centering	
\includegraphics[width=17.5cm,height=6cm ]{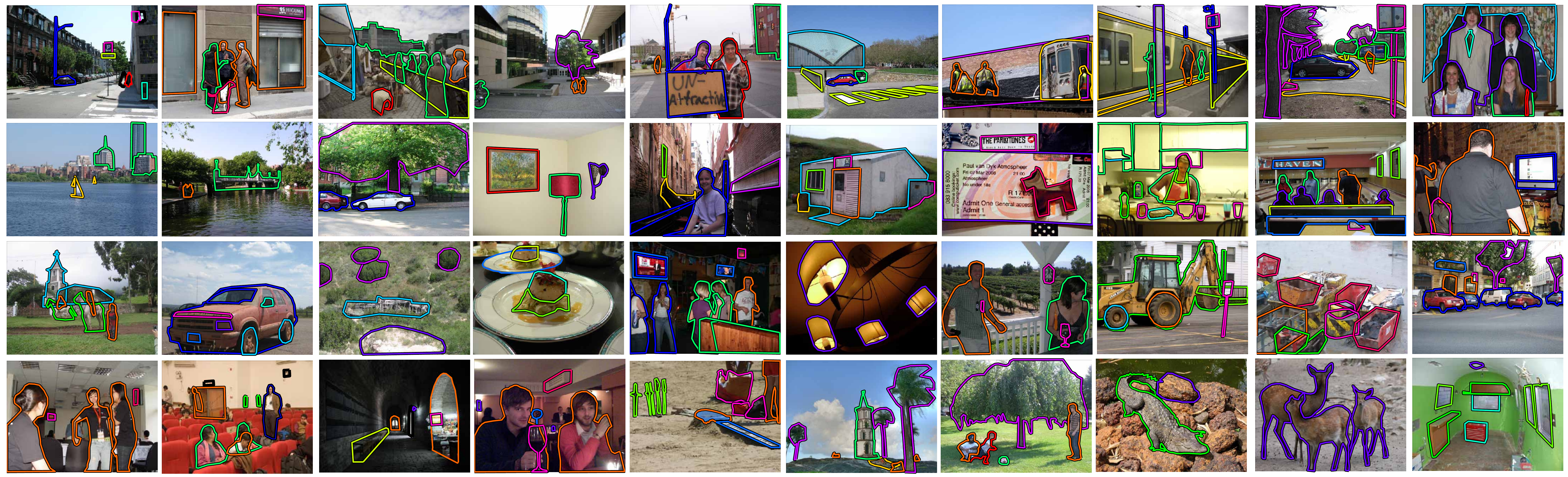} 
\caption{Sample scenes from our dataset and their corresponding annotations.}
  
\label{fig:1}
\end{figure*}

%


We asked 2 observers to manually outline objects using the LabelMe~\cite{LabelMeRef} open annotation
tool (http://new-labelme.csail.mit.edu/). Observers were instructed to accurately segment as
many objects as possible following three rules: 1) discard reflection of objects in mirrors,
2) segment objects that are not separable as one (e.g., apples in a basket), and 3) interpolate
the boundary of occluded objects only if doing otherwise may create several parts for an
occluded object. These cases, however, did happen rarely. Observers were also told that
their outline should be good enough for somebody to recognize the object just by seeing the
drawn polygon. Observers were paid for their effort. Fig.~\ref{fig:1} shows sample images and their
annotated objects. To determine which object is the most salient one, we selected the object
at the peak of the human fixation map.






\begin{figure*}[htp!]
  
\centering	
\includegraphics[width=17cm,height=5.5cm ]{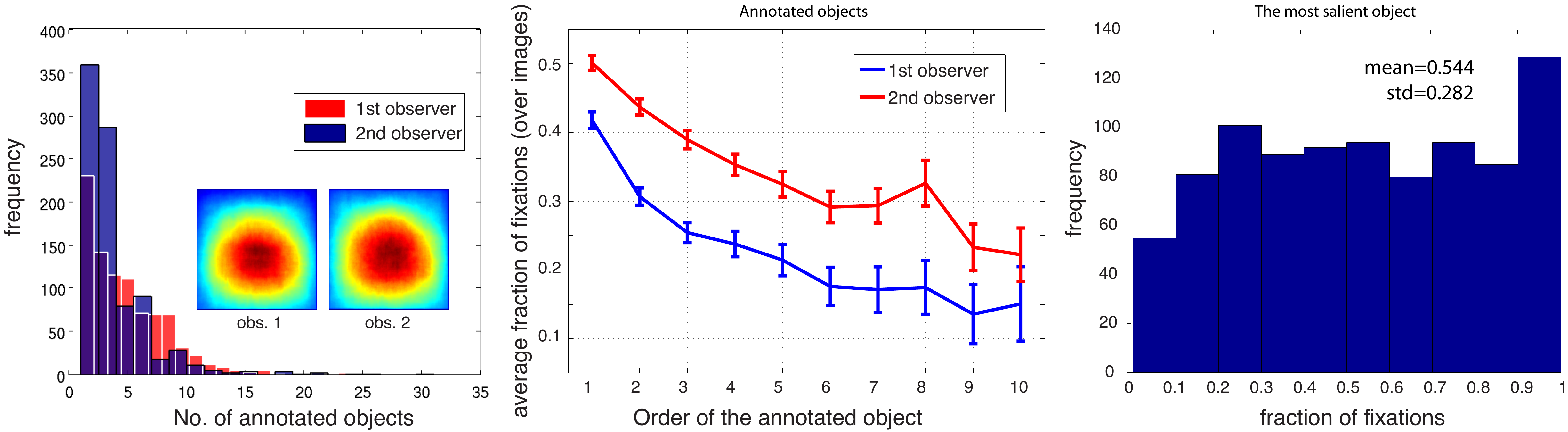} 
\caption{Left: The histogram of number of annotated objects by two observers, Middle: Average fraction of fixations as a function of annotation order. Error bars show the standard error of the mean (s.e.m) over 900 images. 
Right: Histogram of fixation ratios on the most salient object of the Judd-A dataset.}

\label{fig:stat}
\end{figure*}





\subsection{Dataset statistics}
\label{statistics}

Here we explore some summary statistics of our data. 
On average, 36.93\% of an image pixels was annotated by the 1st observer with a std of 29.33\% (44.52\%, std=29.36\% for the 2nd observer). 27.33\% of images 
had more than 50\% of their pixels segmented by the 1st observer (34.18\% for the 2nd). 
The number of annotated objects in a scene ranged from 1 to 31 with median of 3 for the 1st observer (1 to 24 for the 2nd observer with median of 4; Fig.~\ref{fig:stat}.left). 
The median object size was 10\% of the total image area for the 1st observer (9\% for the 2nd observer).
Fig.~\ref{fig:stat}.left (inset) shows the average annotation map for each observer over all images. It indicates that either more objects were present at the image center and/or observers tended to 
annotate central objects more. Overall, our data suggests that both observers agree to a good extent with each other. 
Finally, in order to create one ground truth segmentation map per image, we asked 5 other observers to choose the best of two annotations (criteria based on selection of annotated objects and boundary accuracy). The best annotation was the one with max number of votes (611 images with 4 to 1 votes).


Next, we quantitatively analyzed the relationship between fixations and annotations (Note that we explicitly define the most salient object as the one with the highest fraction of fixations on it). 
We first looked into the relationship between the object annotation order and the fraction of fixations on objects. Fig.~\ref{fig:stat}.middle shows fraction of fixations as a function of object annotation order. In alignment with previous findings~\cite{Elazary_Itti08jov,borji2013stands} we observe that observers chose to annotate objects that attract more fixations. But here, unlike~\cite{Elazary_Itti08jov} which used saliency models to demonstrate that observers prioritize annotating interesting and salient objects, we used actual eye movement data. We also quantized the fraction of fixations that fall on scene objects over the Judd-A annotations, and observed that in about 55\% of images, the most salient object attracts more than 50\% of fixations (mean fixation ratio of 0.54; image background=0.45; Fig.~\ref{fig:stat}.right).

The most salient object ranged in size from 0.1\% to 90.2\% of the image size (median=10.17\%). 
The min and max aspect ratio (W/H) of bounding boxes fitted to the most salient object were 0.04 and 13.7, respectively (median=0.94).


%

\begin{figure}[t] 
  
\centering	
\includegraphics[width=7.5cm,height=4.5cm ]{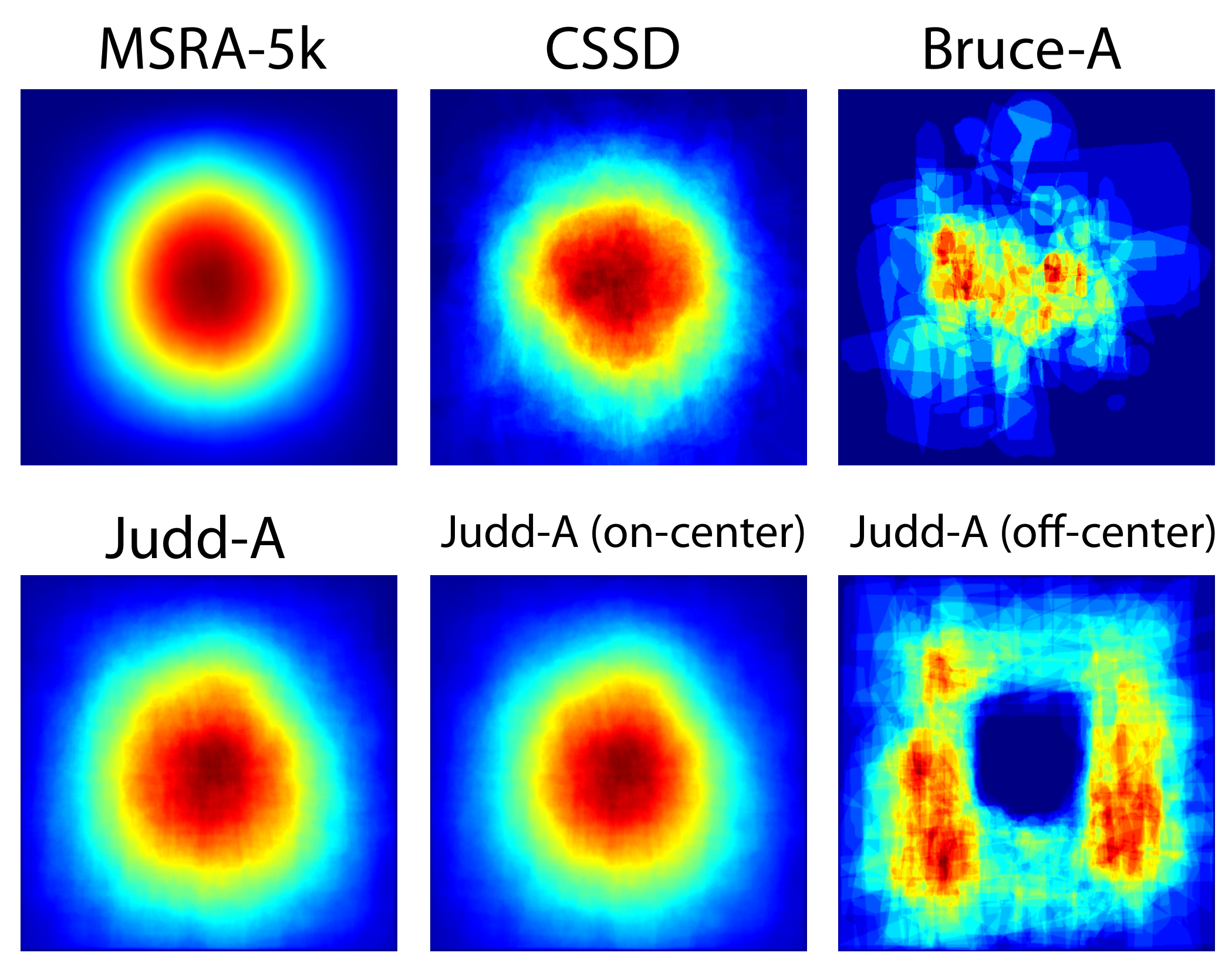} 
\caption{Average annotation maps of two salient object datasets and our datasets.
Distributions of salient objects for on-center and off-center scenes are also shown.}

\label{fig:meps}
\end{figure}

Judd dataset is known to be highly center-biased~\cite{Judd2009,BorjiTIP}, in terms of eye movements~\cite{CenterBiasedTatler}, due to two factors: 1) the tendency of observers to start viewing the image from the center (a.k.a viewing strategy), and 2) tendency of photographers to frame interesting objects at the image center (a.k.a photographer bias). Here we verify the second factor by showing the average annotation map of the most salient object in Fig.~\ref{fig:meps}. Our datasets seem to have relatively less center-bias compared to MSRA-5K and CSSD datasets. Note that other datasets mentioned in Table I are also highly center-biased. To count the number of images with salient objects at the image center, we defined the following criterion. An image is on-centered if its most salient object overlaps with a normalized (to [0 1]) central Gaussian filter with $\sigma=50$. This Gaussian filter is resized to the image size and is then truncated above 0.95. Utilizing this criterion, we selected 667 and 223 on-centered and off-centered scenes, respectively. Partitioning data in this manner helps scrutinize performance of models and tackle the problem of center-bias.

\begin{figure}[htbp!]  
 
\centering	
\includegraphics[width=8.6cm,height=3.6cm ]{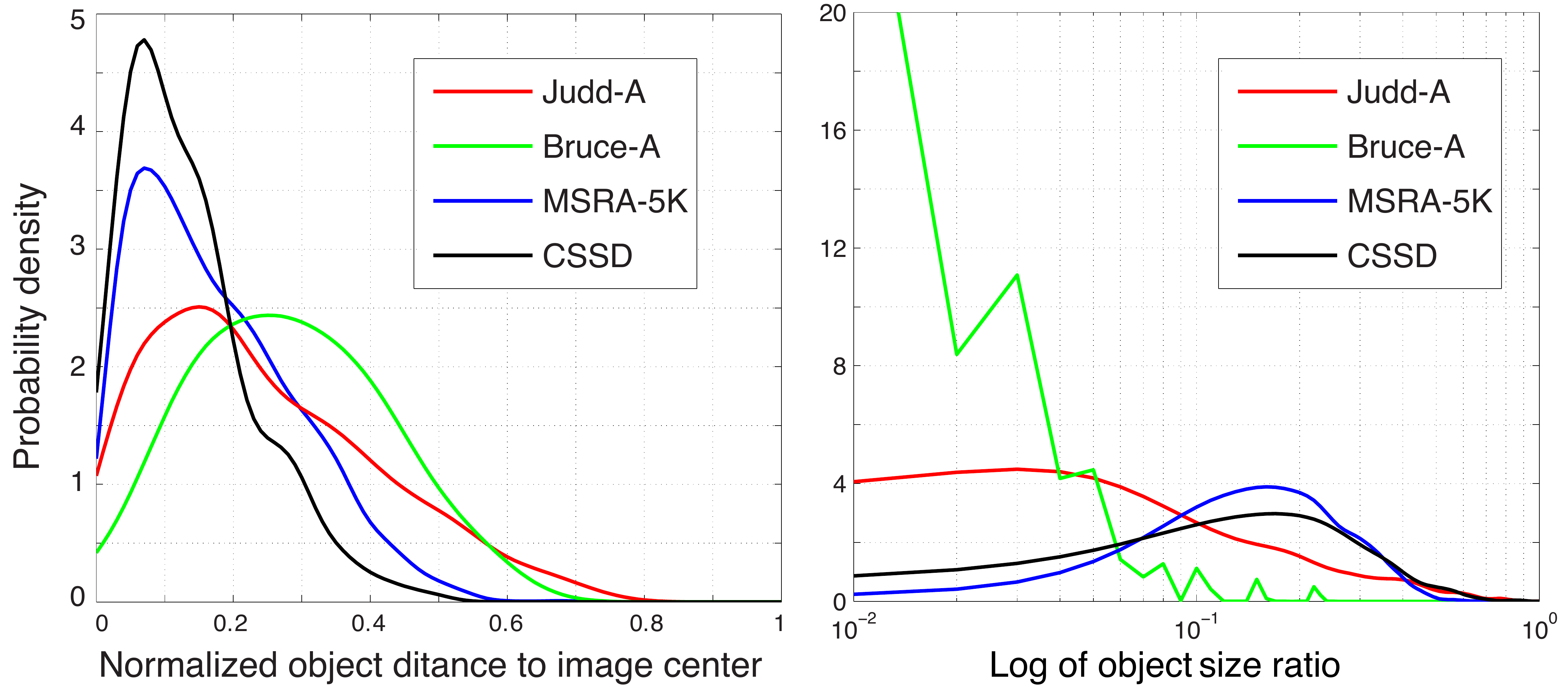} 
\caption{Left: distribution of normalized object distance (\textit{object center - image center} over \textit{half of the image diagonal}). Right: distribution of salient object size ratio (\textit{object size} over \textit{image area}) in log scale across four datasets.}
 
\label{fig:areaRatio}
\end{figure}

To further explore the amount of center-bias in Bruce-A and Judd-A datasets, we first calculated the Euclidean distance from center of bounding boxes, fitted to object masks, to the image center.
We then normalized this distance to the half of the image diagonal (i.e., image corner to image center). Fig.~\ref{fig:areaRatio}.left shows the distribution of normalized object distances.
As opposed to MSRA-5K and CSSD datasets that show an unusual peak around the image center, objects in our datasets are further apart from the image center.

Fig.~\ref{fig:areaRatio}.right shows distributions of normalized object sizes. A majority of salient objects in Bruce-A and Judd-A datasets occupy less than 10\% of the image. On average,
objects in our datasets are smaller than MSRA-5K and CSSD making salient object detection more challenging. 

\begin{figure}[htp!] 

\centering	
\includegraphics[width=8.5cm,height=4.7cm ]{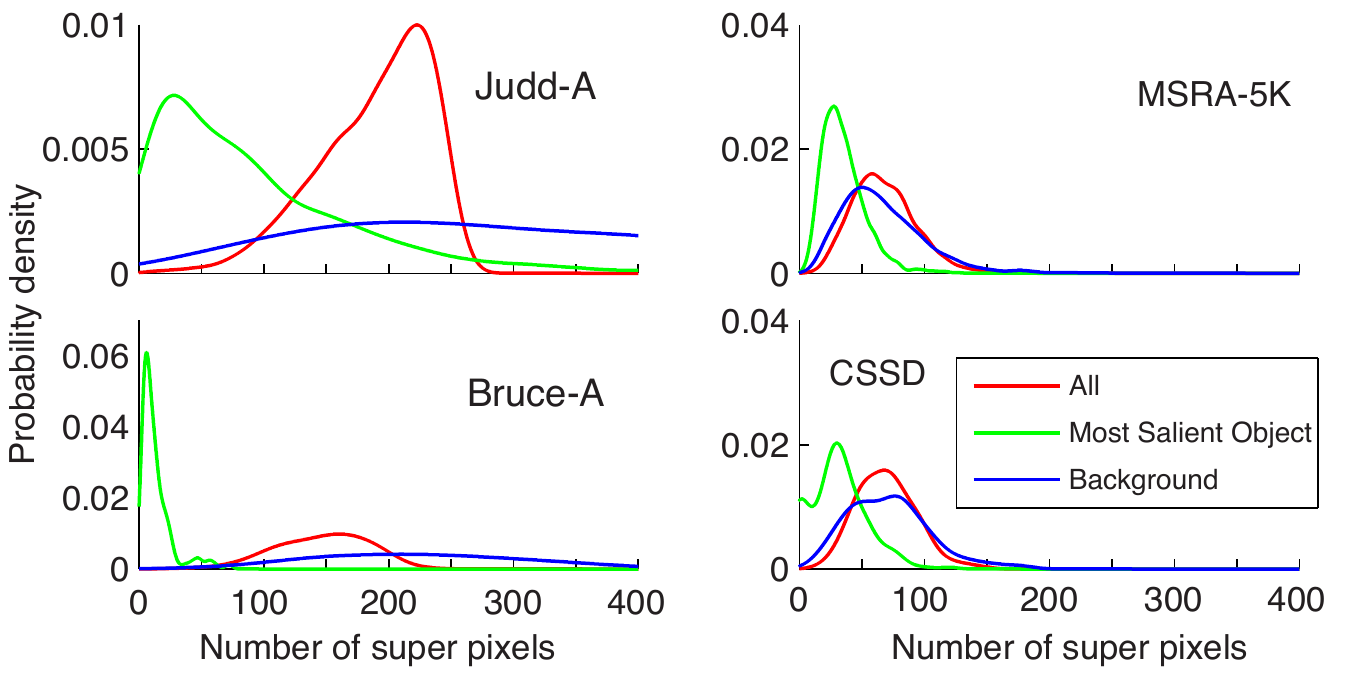} 
\caption{Distributions of the number of superpixels for the most salient object, the background, and entire scene (All) across four datasets.}
 
\label{fig:stats}
\end{figure}

We also analyzed complexity of scenes on four datasets. To this end, we first used the popular graph-based superpixel segmentation algorithm by Felzenszwalb and Huttenlocher~\cite{Felzenszwalb}
to segment an image into contiguous regions larger than 60 pixels each (parameter settings: $\sigma$ = 1, segmentation coefficient $K$ = 300). The basic idea is that the more superpixels an image contains, the more complex and cluttered it is~\cite{Bravo}. By analogy to scenes, an object with several superpixels is less homogeneous, and hence is more complex (e.g., a person vs. a ball
). Fig.~\ref{fig:stats} shows distributions of number of superpixels on the most salient object, the background, and the entire scene. If a superpixel overlapped with the salient object and background, we counted it for both. In general, complexities of backgrounds and whole scenes in our datasets, represented by blue and red curves, 
 are much higher than in the other two datasets. The most salient object in Judd-A dataset on average contains more superpixels than salient objects in MSRA-5K and CSSD datasets, even with smaller objects. The reason why number of superpixels is low on the Bruce-A dataset is because of its very small salient objects (See Fig.~\ref{fig:stats}.right). 


%
%

Further, we inspected types of objects in Judd-A images. We found that 45\% of images have at least one person in them and 27.2\% have more than two people.
On average each scene has 1.56 persons (std = 3.2). In about 27\% of images, annotators chose a person as the most salient object.
We also found that 280 out of 900 images  (31.1\%) had one or more text in them. Other frequent objects were animals, cars, faces, flowers, and signs.



\section{Our baseline saliency model: SalBase}


In general, it is agreed that for good saliency detection, a model should meet the following three criteria: 1) \emph{High detection rate}.
There should be a low probability of failing to detect real salient regions, and low probability of falsely detecting background regions
as salient regions, 2) \emph{High resolution}. Saliency maps should have high or full resolution to accurately locate salient objects and retain original image information as much
as possible, and 3) \emph{High computational efficiency}. Saliency models with low processing time are preferred.
Here, we analyze these factors by proposing a simple baseline salient object detection model.

We propose a straightforward model to serve two purposes: 1) \textit{to assess the degree to which our data can be explained by a simple model}. This way our model can be used for measuring 
bias and complexity of a saliency dataset, and 2) \textit{to gauge progress and performance of the state of the art models}. By comparing performance of best models relative to this baseline model over existing datasets and our datasets, we can judge how powerful and scalable these models are. Note that we deliberately keep the model simple to achieve above goals. 


\begin{figure*}[htbp!]

\centering	
\includegraphics[width=\linewidth]{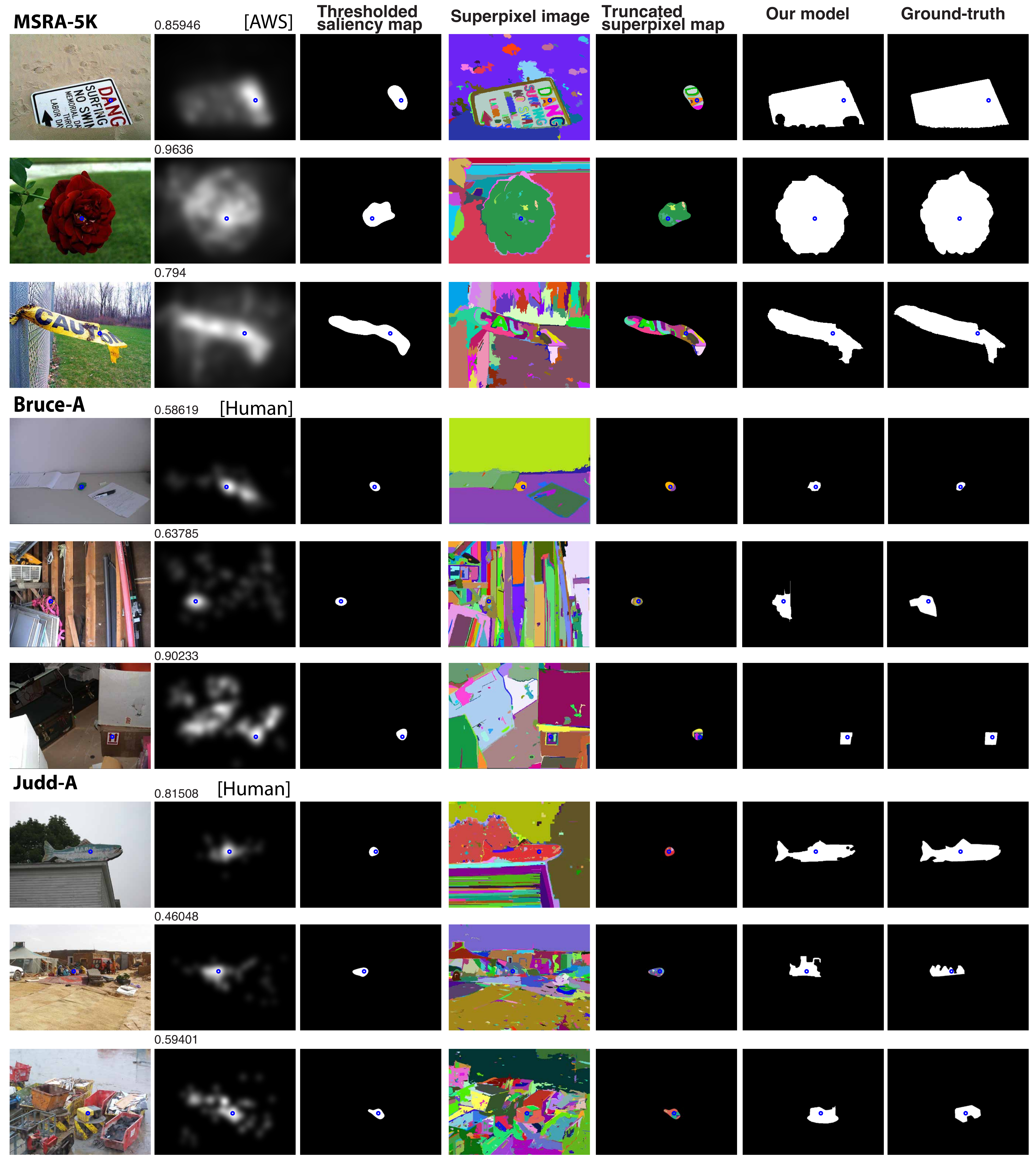}    
\caption{Columns from left to right: original image, saliency map (human or a model), top 70\% of the saliency map ($\beta$ = 0.7), graph-based superpixel segmentation~\cite{Felzenszwalb}, truncated superpixel image, our model prediction, and ground-truth. Small blue dot represents the location of saliency map maximum. Numbers above images in the second column show the PASCAL criterion $\Omega = |\omega \cap o| / |\omega \cup o|$ where $\omega$ is our segmentation and $o$ is the ground-truth annotation, so the higher the $\Omega$ the better.}
 
\label{fig:samples}
\end{figure*}

Our model involves the following two steps:\\ 
\noindent {\em \textbf{Step 1}}: Given an input image, we compute a saliency map and an over-segmented region map. 
For the former, we use a fixation prediction model (traditional saliency models) to find spatial outliers in scenes that attract human eye movements and visual attention. 
Here, we use two models for this purpose: AWS~\cite{GarciaDiaz} and HouNIPS~\cite{HouZhangNIPS2008}, which have been shown to perform very well in recent benchmarks and to be computationally efficient~\cite{BorjiTIP}.
As controls, we also use the generic \textit{objectness} measure by Alexe {\em et al.}~\cite{Alexe}, as well as the human fixation map to determine the upper-bound performance. The reason for using fixation saliency models is to 
obtain an quick initial estimation of locations where people may look in the hope of finding the most salient object. These regions are then fed to the segmentation component in the next step. It is critical to first limit the subsequent expensive processes onto the right region. For the latter, as in the previous section, we use the fast and robust algorithm by Felzenszwalb and Huttenlocher~\cite{Felzenszwalb}\footnote{We achieved lower accuracy using the normalized cut algorithm~\cite{Shi}.} with same parameters as in section~\ref{statistics}.

\begin{figure*}[htp!]

\centering	
\includegraphics[width=18.2cm,height=9.8cm ]{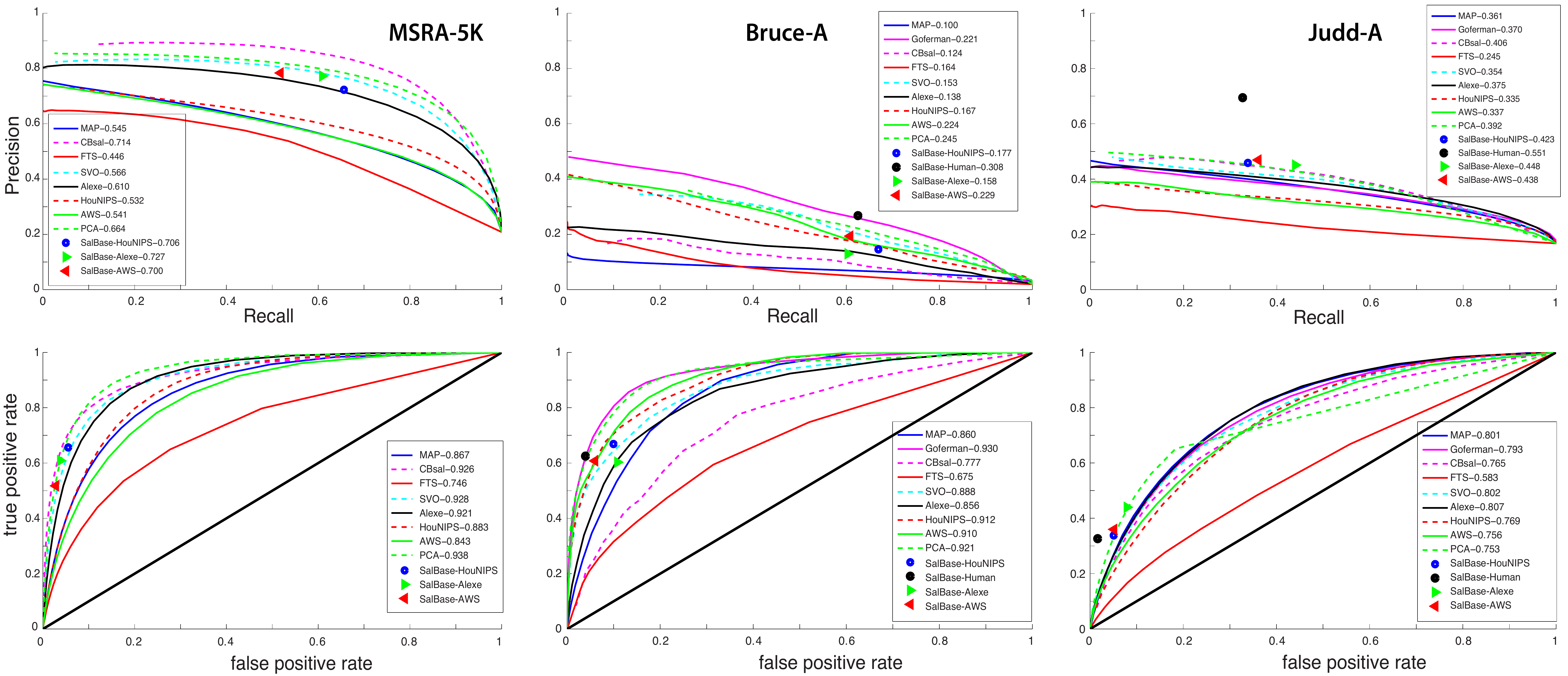} 
\caption{PR and ROC curves of our model versus 8 other models. 
Note how drastically models are degraded on our dataset, which contain scenes with multiple objects. We did not run Goferman model~\cite{Goferman} over MSRA-5K due to its slow computation. Alexe model~\cite{Alexe} is used with 1000 windows. MAP stands for the mean annotation map (Fig.~\ref{fig:meps}). Also note that we expect similar performance for even better recent salient object detection models (e.g.,DRFI~\cite{JiangCVPR2013}).}

\label{fig:Bruce}
\end{figure*}


\noindent {\em \textbf{Step 2}}: The saliency map is first normalized to [0 1] and is then thresholded at $\beta$ (here $\beta$ = 0.7).
Then all unique image superpixels that spatially overlap with the truncated saliency map are included.  
Here we discarded those superpixels that touch the image boundary because they are highly likely to be part of the background. Finally, after this process, the holes inside the selected region will be considered as part of the salient object (e.g., filling in operation).
Fig.~\ref{fig:samples} illustrates the process of segmentation and shows outputs of our model for some images from MSRA-5K, Bruce-A, and Judd-A datasets.

The essential feature of our simple model is dissociating saliency detection from segmentation, such that now it is possible to pinpoint what might be the cause of mistakes or low performance of a model, i.e., \underline{detecting the wrong object} or \underline{faulty segmentation}. This is particularly important since almost all models have confused these two steps and have faded the boundary. 

Note that currently there is no training stage in our model and it is manually constructed with fixed parameters.
The second stage in our model is where more modeling contribution can be made, for example by devising more elaborate ways to include or discard superpixels in the final segmentation. One strategy is to learn model parameters from data. Some features to include in a learning method are size and position of a superpixel, a measure of elongatedness, a measure of concavity or convexity, distance between feature distributions of a superpixel and its neighbors, etc. To some extent, some of these these features have already been utilized in previous models~\cite{BorjiECCV,MingCheng}. Another direction will be expanding our model to multi scale (similar to~\cite{Yan}).

\section{Model evaluation and comparison}
 
We exhaustively compared our model to 8 state of the art methods which have been shown to 
perform very well on previous benchmarks~\cite{BorjiECCV}. These models come from 3 categories allowing us to perform cross-category comparison:
1) \textit{salient object detection models} including CBsal~\cite{Jiang}, SVO~\cite{Chang}, PCA~\cite{Margolin}, Goferman~\cite{Goferman}, and FTS~\cite{AchantaCVPR}, 2) \textit{generic objectness measure} by Alexe {\em et al.}~\cite{Alexe}, and 3) \textit{fixation prediction models} including AWS~\cite{GarciaDiaz} and HouNIPS~\cite{HouZhangNIPS2008}.  

We use two widely adopted metrics: 
\begin{itemize}
 \item \textbf{Precision-recall (PR) curve:}
For a saliency map $S$ normalized to $[0, 255]$, we convert it to a binary mask $M$ with a threshold $T_f$. $Precision$ and $Recall$ are then computed as follows given the ground truth mask $G$:
\begin{align}
  Precision = \frac{|M\cap G|}{|M|}, \ \ Recall = \frac{|M\cap G|}{|G|}
\end{align}
\label{eqn:precision_recall}
To measure the quality of saliency maps produced by several algorithms, we vary the threshold $T_f$ from 0 to 255. On each threshold, $Precision$ and $Recall$ values are computed. Finally, we can get a precision-recall (PR) curve to describe the performance of different algorithms.

We also report the F-Measure defined as:
\begin{equation}
 F_{\alpha} = \frac{(1+\alpha) Precision  \times  Recall}{\alpha \times Precision + Recall}
\end{equation}

 Here, as in~\cite{AchantaCVPR} and~\cite{MingCheng}, we set $\alpha = 0.3$ to weigh precision more than recall.

 \item \textbf{Receiver operating characteristics (ROC) curve:} We also report the false positive rate ($FPR$) and true positive rate ($TPR$) during the thresholding a saliency map:
\begin{align}
TPR = \frac{|M\cap G|}{|G|}, \ \ FPR = \frac{|M\cap G|}{|M\cap G| + |\bar{M}\cap\bar{G}|}
\end{align}
where $\bar{M}$ and $\bar{G}$ denote the opposite (complement) of the binary mask $M$ and ground-truth, respectively. The ROC curve is the plot of $TPR$ versus $FPR$ by varying the threshold $T_f$.

\end{itemize}

\subsection{Quantitative evaluation}
\begin{figure}[htp!]
  
\centering	
\fbox{\includegraphics[width=8.5cm,height=5.5cm ]{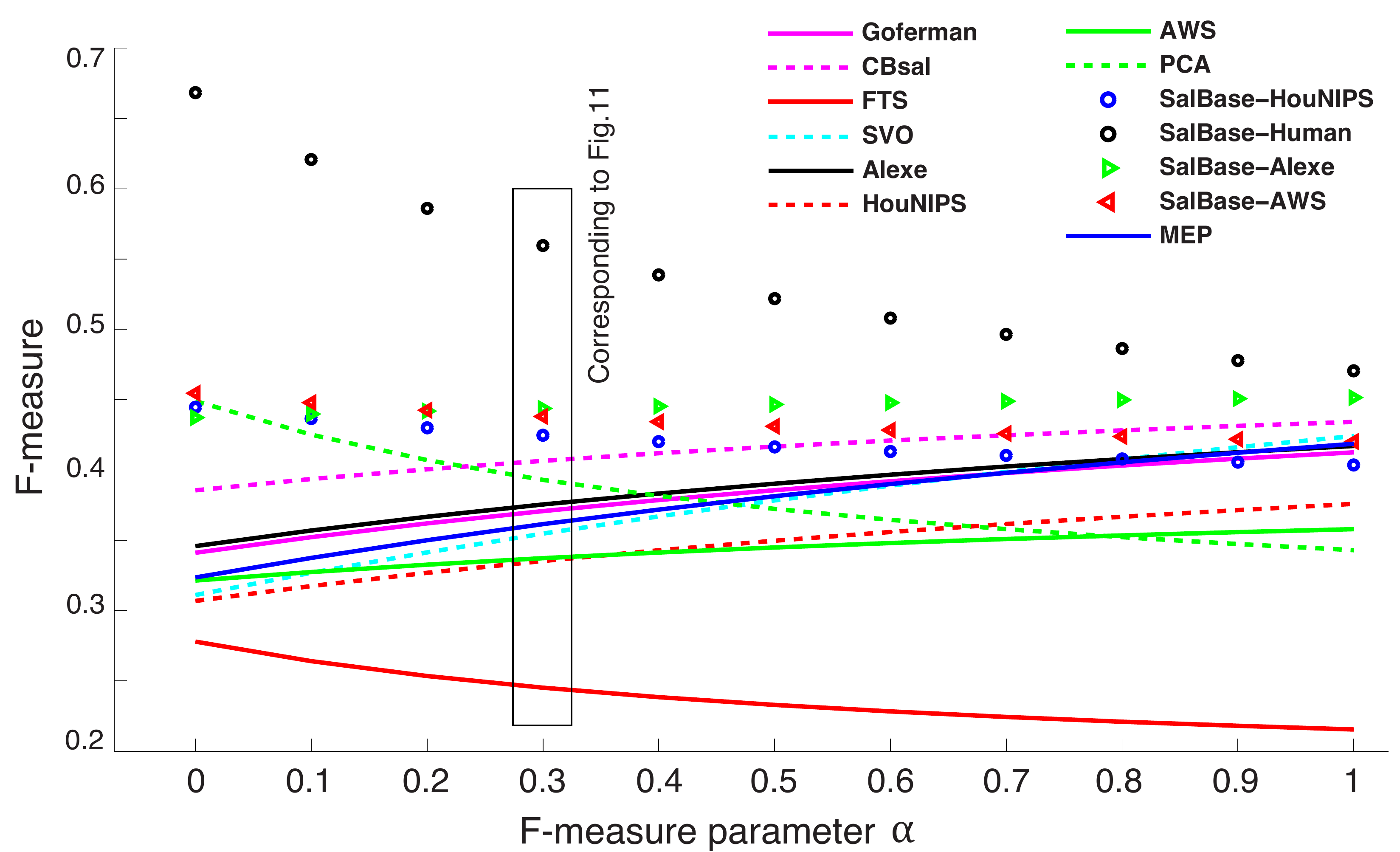}} 
\caption{F-measure as a function of parameter $\alpha$ on Judd-A dataset.}
\label{fig:F-measure2}
\end{figure}

Results are shown in Fig.~\ref{fig:Bruce}. Consistent with previous reports over the MSRA-5K dataset~\cite{JiangCVPR2013}, CBsal, PCA, SVO, and Alexe models rank on the top (with F-measures above 0.55 and AUCs above 0.90).
Fixation prediction models perform lower at the level of the MAP. FTS model ranked on the bottom again in alignment with previous results. Our models work on par with the best models on this dataset
with all F-measures above 0.70 (max with Alexe model about 0.73). Moving from this simple dataset (because our simple models ranked on the top; see also the analysis in section~\ref{statistics}) to more complex datasets (middle column in Fig.~\ref{fig:Bruce}) we observed a dramatic drop in performance of all models. The best performance now is 0.24 belonging to the PCA model. We observed about 72\% drop in performance averaged over 5 models (CBsal, FTS, SVO, PCA, and Alexe) from MSRA-5K to Bruce-A dataset. Note in particular how MAP model is severely degraded here  (poorest with F measure of 0.1) since objects are now less at the center. 
Our best model on this dataset is the SalBase-Human (F-measure about 0.31). Surprisingly, AUC results are still high on this dataset since objects are small thus true positive rate is high at all levels of false positive rate (See also performance of MAP). Patterns of results over Judd-A dataset are similar to those over Bruce-A with all of our models performing higher than others. The lowest performance here belongs to FTS followed by the two fixation prediction models. Our SalBase-Human model scores the best with the F-measure about 0.55. Among our models that used a model to pick the most salient location, SalBase-AWS scores higher over Bruce-A and Judd-A datasets possibly because AWS is better able to find the most salient location. The average drop from MSRA-5K to Judd-A dataset is $\sim$ 41\% (for 5 saliency detection models). Fig.~\ref{fig:F-measure2} shows that these findings are robust to F-measure parameterization.
Tables~\ref{tab:db-Fmeasure} and~\ref{tab:db-AUC} summarize the F-measure and AUC of models.

\begin{table}[t]
\renewcommand{\arraystretch}{1}
\renewcommand{\tabcolsep}{0.5mm}
\begin{center}
\begin{tabular}{|c|c|c|c|}
\hline
\textbf{Model} & \multicolumn{3}{c|}{\textbf{Dataset}}  \\
    \cline{2-4}            &   MSRA - 5K  &  Bruce-A     &  Judd-A   \\ \hline\hline
MEP & 0.545 & 0.10  & 0.361 \\
Goferman & - & 0.221  & 0.370 \\
CBsal & 0.714 & 0.124  & 0.406 \\
FTS & 0.446 & 0.164  & 0.245 \\
SVO & 0.566 & 0.153  & 0.354 \\
Alexe & 0.610 & 0.138  & 0.375 \\
HouNIPS & 0.532 & 0.167  & 0.335 \\
AWS & 0.541 & 0.224  & 0.337 \\
PCA & 0.664 & 0.245  & 0.392 \\
\hline
SalBase-HouNIPS & 0.706 & 0.177  & 0.423\\
SalBase-Alexe & \textbf{0.727} & 0.158  & 0.448 \\
SalBase-AWS & 0.700 & 0.229  & 0.438 \\
SalBase-Human & - & \textbf{0.308}  & \textbf{0.551} \\
\hline
\end{tabular}
\end{center}
\caption{F-measure accuracy of models. Performance of the best model is highlighted in boldface font.}
\label{tab:db-Fmeasure}

\end{table}

\begin{table}[t]
\renewcommand{\arraystretch}{1}
\renewcommand{\tabcolsep}{0.5mm}
\begin{center}
\begin{tabular}{|c|c|c|c|}
\hline
\textbf{Model} & \multicolumn{3}{c|}{\textbf{Dataset}}  \\
    \cline{2-4}            &   MSRA - 5K  &  Bruce-A     &  Judd-A   \\ \hline\hline
MEP & 0.867 & 0.860  & 0.801 \\
Goferman & - & \textbf{0.930}  & 0.793 \\
CBsal & 0.926 & 0.777  & 0.765 \\
FTS & 0.746 & 0.675  & 0.583 \\
SVO & 0.928 & 0.888  & 0.802 \\
Alexe & 0.921 & 0.856  & \textbf{0.807} \\
HouNIPS & 0.883 & 0.912  & 0.769 \\
AWS & 0.843 & 0.910  & 0.756 \\
PCA & \textbf{0.938} & 0.921  & 0.753 \\
\hline
SalBase-HouNIPS & 0.781 & 0.751  & 0.633\\
SalBase-Alexe & 0.773 & 0.714  & 0.662 \\
SalBase-AWS & 0.737 & 0.756  & 0.644 \\
SalBase-Human & - & 0.780  & 0.654 \\
\hline
\end{tabular}
\end{center}
\caption{AUC accuracy of models. Performance of the best model is highlighted in boldface font.}
\label{tab:db-AUC}

\end{table}

\subsection{Analysis of saliency map thresholding}
\begin{figure}[htp!]
\centering	
\includegraphics[width=8.5cm,height=3.8cm ]{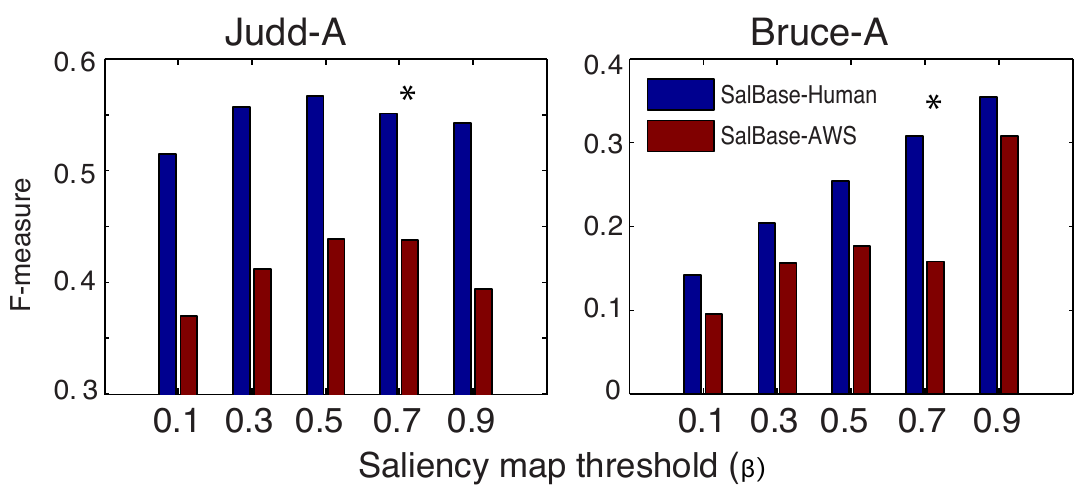} 
\caption{F-measure as a function of saliency threshold $\beta$. Stars correspond to points shown in Fig.~\ref{fig:Bruce}. Note that even higher accuracies are possible with different thresholds over our datasets. Corresponding F-measure values over MSRA-5K for SalBase-AWS model are: 0.62, 0.67, 0.70, 0.73, and 0.62.}
 \label{fig:salThreshold}
\end{figure}

To study the dependency of results on saliency map thresholding (i.e., how many superpixels to include), we varied the saliency threshold $\beta$ and calculated F-measure for SalBase-Human and SalBase-AWS models (See Fig.~\ref{fig:salThreshold}). We observed that even higher scores are achievable using different parameters. For example, since objects in the Judd-A dataset are larger, a lower threshold yields a better accuracy. The opposite holds over the Bruce-A dataset.

\subsection{Analysis of superpixel segmentation parameters}
\begin{figure*}[t]
  
\centering	
\includegraphics[width=17.7cm,height=4.2cm ]{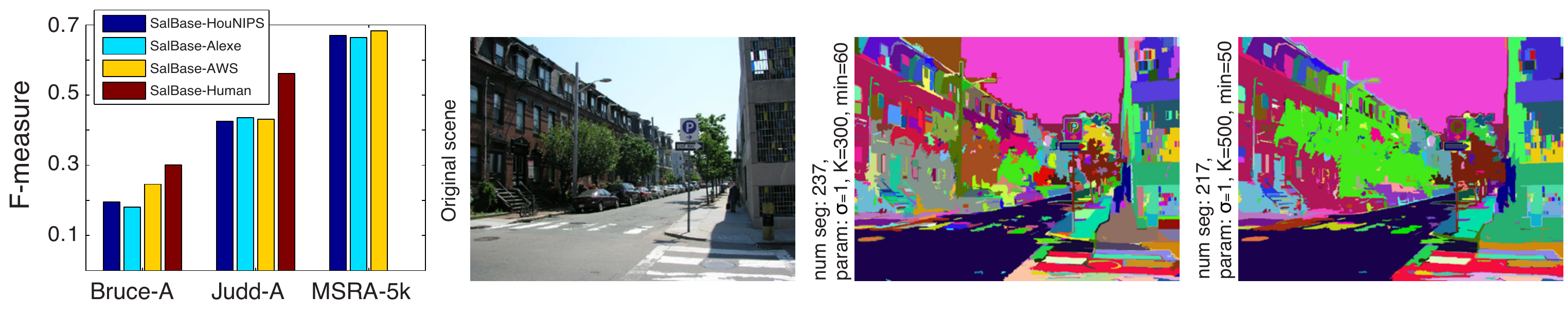} 
\caption{Left: F-measure with superpixel parameters as $\sigma$ = 1, K = 500, and min = 50. Right: A sample image and its corresponding superpixel segmentations.}
\label{fig:res_500_50}
\end{figure*}

To investigate the dependency of results on segmentation parameters, we varied the parameters of the segmentation algorithm from too fine ($\sigma$ = 1, K = 100, min = 20; many segments; over-segmenting) to too coarse ($\sigma$ = 1, K = 1000, min =  800; fewer segments; under-segmenting). Both of these settings yielded lower performances than results in Fig.~\ref{fig:Bruce}. Results with another parameter setting with $\sigma$ = 1, K = 500, and min = 50 are shown in 
Fig.~\ref{fig:res_500_50}. Scores and trends are similar to those shown in Fig.~\ref{fig:Bruce}, with SalBase-Human and SalBase-AWS being the top contenders.

\subsection{Analysis of model failure cases}
\begin{figure*}[t]

\centering	
\includegraphics[width=.9\linewidth]{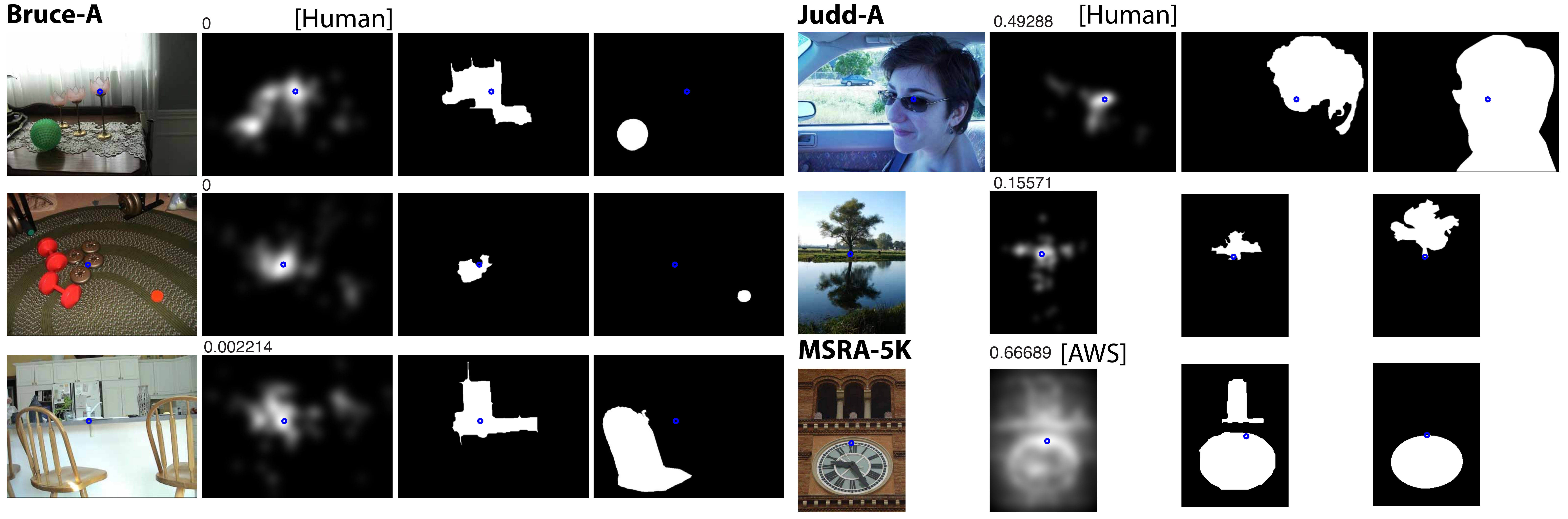} 
\caption{Sample images where our model (3rd column) fails. Fourth columns show the ground truth. Left: images from the Bruce dataset where our model fails because 
human saliency judgments and fixations do not agree. Right: failure cases where some superpixels are mistakenly discarded or included.}
 
\label{fig:failure2}
\end{figure*}

Analysis of cases where our model fails, shown in Fig.~\ref{fig:failure2}, reveals four reasons: \textit{First}, on Bruce-A dataset when humans look at an object more but annotators chose a different object. 
\textit{Second}, when a segment that touches the image border is part of the salient object. \textit{Third}, when the object segment falls outside the thresholded saliency map (or a wrong one is included). \textit{Fourth}, when the first stage 
(i.e., fixation prediction model) pick the wrong object as the most salient one (See Fig.~\ref{fig:rrr}, first column). Regarding the first problem, care must be taken in assuming what people look is what they choose as the most salient object.
Although this assumption is correct in a majority of cases (Fig.~\ref{fig:exp1}), it does not hold in some cases. With respect to the second and third problems, future modeling effort is needed to decide which superpixels to include/discard to 
determine the extent of an object. The fourth problem points toward shortcomings of fixation prediction models. Indeed, in several scenes where our model failed, people and text were the most salient objects. 
Person and text detectors were not utilized in the saliency models employed here. 

\subsection{Qualitative comparison of models}
\begin{figure*}[!htp]
  
\centering	
\includegraphics[width=18cm,height=17cm ]{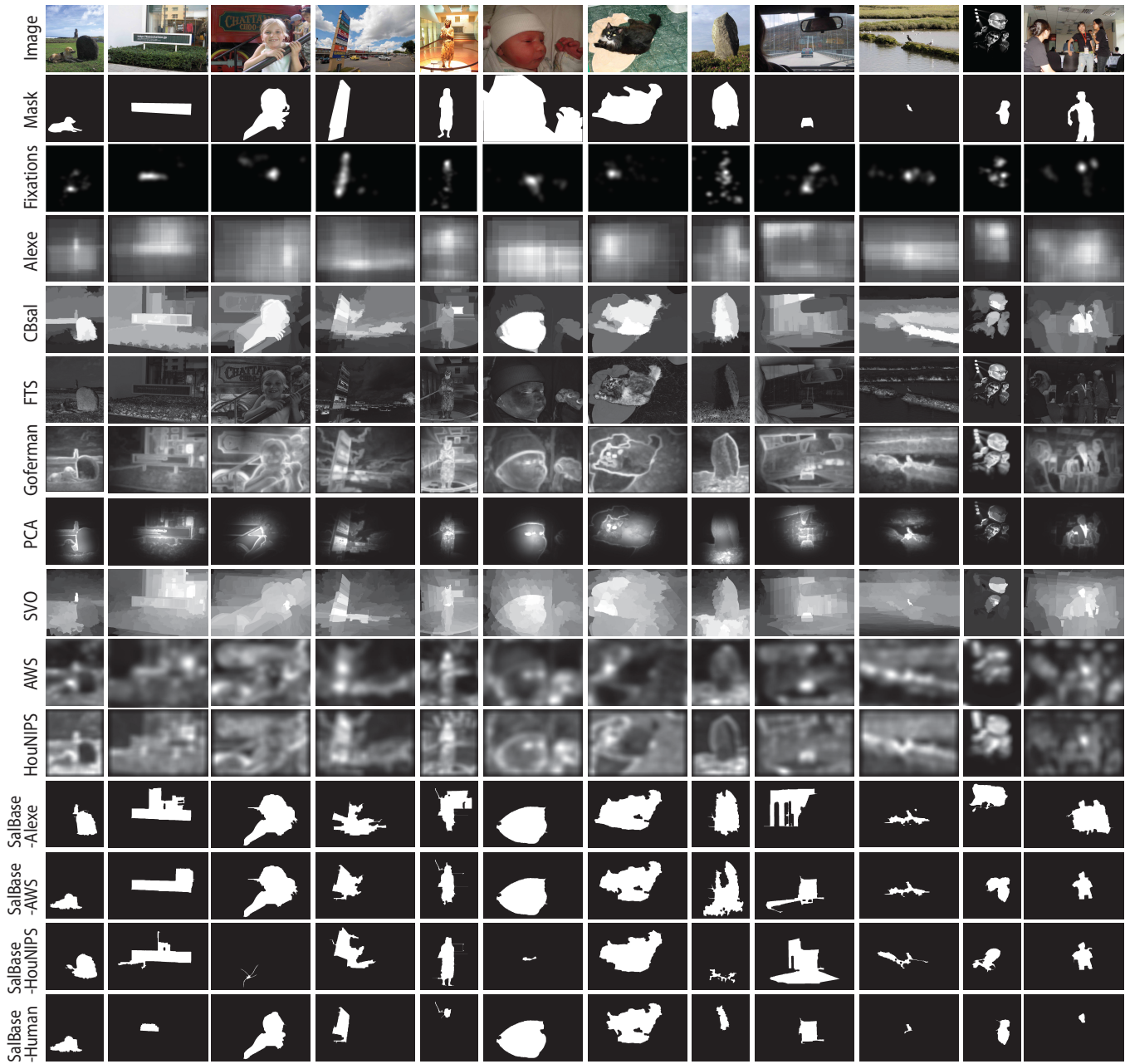}
\caption{\footnotesize{Sample images from Judd-A dataset along with saliency detection maps of models.}}
 
\label{fig:rrr}
\end{figure*}

Fig.~\ref{fig:rrr} shows a visual comparison of models over 12 scenes from the Judd-A dataset. CBsal and SVO generate more visually pleasant maps. Goferman highlights object boundaries more than 
object interiors. PCA generates center-biased maps. Some models (e.g., Goferman, FTS) generate sparse saliency maps while some others generate smoother ones (e.g., SVO, CBSal). AWS and HouNIPS models generate pointy maps to better account for fixation locations.

\section{Discussion and Conclusion}
\label{conclusions}
In this work, we showed that: 1) explicit human saliency judgments agree with free-viewing fixations (thus extending our previous results in~\cite{borji2013stands}), 2) our new 
benchmark datasets challenge existing state-of-the-art salient object detection models (in alignment with Li~{\em et al.}'s dataset~\cite{YiHou2014}), and 3) a conceptually simple and computationally efficient model ($\sim$0.2 s for $400\times300$ saliency and segmentation maps on a PC with a 3.2 GHz Intel i7 CPU and 6GB RAM using Matlab) wins over the state of the art models and can be used as a baseline in the future. We also highlighted a limitation of models which is the main reason behind their failure on complex scenes. They often segment the wrong object as the most salient one.


Previous modeling effort has been mainly concentrated on biased datasets with images containing objects at the center. Here, we focused on this shortcoming and described how unbiased salient object detection datasets can be constructed. We also reviewed datasets that can be used for saliency model evaluation (in addition to datasets in Table~\ref{tab:db}) and measured their statistics. No dataset exists so far that has all of object annotations, eye movements, and explicit saliency judgments. Bruce-A has fixations, and only explicit
saliency judgments but not all object labels. Judd-A, OSIE, and PASCAL-S datasets have annotations and fixations but not explicit saliency judgments. Here, we chose the object that falls at the peak of the fixation map as the most salient one. UCSB dataset lacks object annotations but it has fixations and saliency judgments using clicks (as opposed to object boundaries in Bruce-A). 
Future research by collecting all information on a large scale dataset will benefit salient object detection research.

Here we suggested that the most salient object in a scene is the one that attracts the majority of fixations (similar to~\cite{YiHou2014}). One can argue that the most salient object is the one that observers look at first. While in general, these two definitions may choose different objects, given the short presentation times in our datasets (3 sec on Judd, 4 sec on Bruce) we suspect that both suggestions will yield
to similar results.

Our model separates detection from segmentation. A benefit of this way of modeling is that it can be utilized for other purposes (e.g., segmenting interesting or important objects) by replacing
the first component of our model. Further, augmented with a top-down fixation selection strategy, our model can be used as an active observer (e.g.,~\cite{borjiSMC}).

Our analysis suggests two main reasons for model performance drop over the Judd-A dataset: The \textit{first reason} that the literature has focused so far is to avoid incorrectly segmenting the object region (i.e., increasing true positives and reducing false positives). Therefore, low performance is partially due to inaccurately highlighting (segmenting) the salient object. 
The \textit{second reason} that we attempted to highlight in this paper (we believe is the main problem causing performance drop as models performed poorly on Judd-A compared to MSRA-5K) is segmenting the wrong object (i.e., not the most salient object).
Note that although here we did not consider the latest proposed salient object detection models in our model comparison (e.g.,~\cite{yang2013saliency,Perazzi,Jiang1,Jiang2,JiangCVPR2013,Kim}), we believe that our results are likely to generalize compared to newer models. The rationale is that even recent models have also used the ASD dataset~\cite{AchantaCVPR} (which is highly center-biased) for model development and testing. Nontheless, we encourage future works to use our model (as well as Li et al.'s model) as a baseline for model benchmarking.

Two types of cues can be utilized for segmenting an object: appearance~\cite{Felzenszwalb,Shi} (i.e., grouping contiguous areas based on surface similarities) and boundary~\cite{Martin_etal04pami,Ren_etal05iccv} (i.e., cut regions based on observed pixel boundaries). Here we mainly focused on the appearance features. Taking advantage of both region appearance and contour information (similar to~\cite{Arbelaez_etal11pami,Mishra}) for saliency detection (e.g., growing the foreground salient region until reaching the object boundary) is an interesting future direction. In this regard, it will be helpful to design suitable measures for evaluating accuracy of models for detecting boundary (e.g.,~\cite{Movahhedi}).

Our datasets allow more elaborate analysis of the interplay between saliency detection, fixation prediction, and object proposal generation. Obviously, these models depend on the other.
On one hand, it is critical to correctly predict where people look to know which object is the most salient one. On the other hand, labeled objects in scenes can help us study how objects guide attention and eye movements. For example, by verifying the hypotheses that some parts of objects (e.g., object center~\cite{Nuthman2010}) or semantically similar objects~\cite{HwangPomplun2011}) attract fixations more, better fixation prediction models become feasible.




\begin{biography}[{\includegraphics[width=1in,height=1.25in,clip,keepaspectratio]{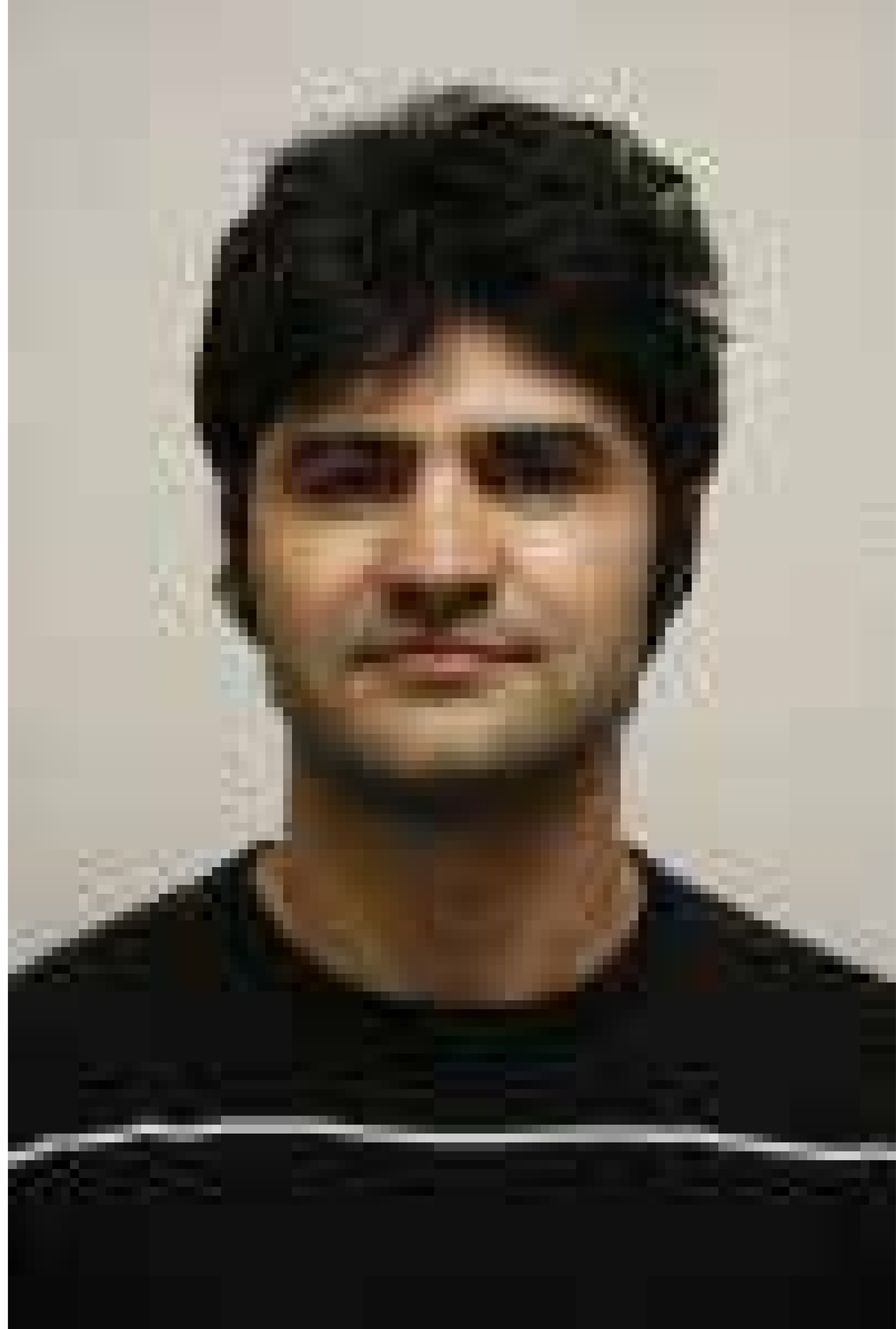}}]{Ali Borji}
received his BS and MS degrees in computer engineering from Petroleum University of Technology, Tehran, Iran, 2001 and Shiraz University, Shiraz, Iran, 2004, respectively. He did his Ph.D. in cognitive neurosciences at Institute for Studies in Fundamental Sciences (IPM) in Tehran, Iran, 2009 and spent four years as a postdoctoral scholar at iLab, University of Southern California from 2010 to 2014. He is currently an assistant professor at University of Wisconsin, Milwaukee. His research interests include visual attention, active learning, object and scene recognition, and cognitive and computational neurosciences.
\end{biography}


\end{document}